\documentclass[12pt]{article}

\usepackage{sbc-template}
\usepackage{hyperref}
\usepackage{url}
\usepackage{booktabs}
\usepackage[english]{babel}   
\usepackage[utf8]{inputenc}  
\usepackage[pdftex]{color,graphicx}
\usepackage{comment}

\usepackage{amsfonts,dcolumn}

\hypersetup{
    bookmarks=true,         
    unicode=true,          
    pdftoolbar=true,        
    pdfmenubar=true,        
    pdffitwindow=true,     
    pdfcreator={},   
    pdfproducer={},  
    pdfkeywords={}, 
    pdfnewwindow=true,      
    colorlinks=true,       
    linkcolor=red,          
    citecolor=green,        
    filecolor=magenta,      
    urlcolor=cyan           
}

\title{An Automatic Patch-based Approach for HER-2 Scoring in Immunohistochemical Breast Cancer Images Using Color Features}

\author{Caroline Quadros Cordeiro\inst{1}, Sergio Ossamu Ioshii\inst{2}, Jeovane Honório Alves\inst{1}, \\Lucas Ferrari de Oliveira\inst{1}}

\address{Universidade Federal do Paraná (UFPR)\\
  Laboratório de Visão, Robótica e Imagem (VRI)\\
  Curitiba - PR, Brasil
\nextinstitute
  Pontifícia Universidade Católica do Paraná (PUCPR)\\
  Pós-Graduação de Tecnologia em Saúde da PUCPR\\
  Curitiba - PR, Brasil
  \email{cqcordeiro@inf.ufpr.br, sergio.ioshii@pucpr.br} 
  \email{\{jhalves, lferrari\}@inf.ufpr.br}
}

\begin{document} 

\maketitle

\begin{abstract}
Breast cancer (BC) is the most common cancer among women worldwide, approximately 20-25\% of BCs are HER-2 positive. Analysis of HER-2 is fundamental to defining  the appropriate therapy for patients with breast cancer. Inter-pathologist variability in the test results can affect diagnostic accuracy. The present  study intends to propose an automatic scoring HER-2 algorithm. Based on color  features, the technique is fully-automated and avoids segmentation, showing a concordance higher than 90\% with a pathologist in the experiments realized. 
\end{abstract}

\section{Introduction}
\label{sec:intro}


Cancer is a disease with a high mortality rate that has increasingly reached the world’s population, especially the female population. 
In Brazil, Breast Cancer (BC) is the most common tumor among women, affecting almost 60,000 patients in 2014 \cite{inca14}.
BC is the second most common tumor worldwide. In US one out of eight women are affected by BC during their lifetime \cite{desantis14}.
In the last decade, incidence of cancer has grown 20\% in the world. In Brazil, \textit{National Institute of Cancer José Alencar Gomes da Silva} (INCA) estimates 59,700 new BC cases in 2018 \cite{inca18}. According to the International Agency for Research on Cancer (IARC), while cancer mortality rate increased by 8\% in 2012, mortality rate of BC was 14\% in the same period \cite{jaques15}.

In breast cancer patients, the amplification of the \textit{Her2} (Human Epidermal growth factor Receptor-type 2) gene is an individual prognosticator and a predictive marker of response to targeted treatment with trastuzumab and adjuvant chemotherapy \cite{slamon01}. Approximately 20-25\% of BCs are HER-2 positive \cite{yaziji04}.


For HER-2 score determination, immunohistochemical tests (IHC) are performed. HER-2 test indicates whether this protein is carrying some role in the development of breast cancer, since with many HER-2 receptors, the cells receive many signals to grow and split. The amount of HER-2 is scored as 0, 1+, 2+ or 3+. If the score is 0 or 1+, it is called "HER-2 negative"; if the score is 2+, then it is called "limit"; and a 3+ score is called "HER-2 positive" \cite{kumar13}. 

HER-2 scoring still has a visual and manual analysis of histological tissues as a standard method. Such method is strongly dependent on the expertise and experience of histopathologists and has the disadvantages of being time-consuming and non-replicable \cite{aktan16}. Some HER-2 tests may present different results, indicating the existence of variations within and between specialist observation \cite{kumar13}.

In the past few years, several works were developed for HER-2-assisted computer classification. Most are commercial, depend on specific materials and are financially costly \cite{brugmann12, jeung12, dobson10, viale16}. Also, methods developed in other works did not show much agreement with pathologists \cite{aktan16, tuominen12, skaland08, masmoudi09, hall08, joshi07}. Currently, some HER-2 scoring software are available on the market. Among then are the Automated Cellular Imaging System III (ACIS III) (Dako) and the HER2-CONNECT (Visiopharm). 

The Automated Cellular Imaging System III (ACIS III) (Dako) was evaluate about correlation between manual HER-2 scoring and HER-2 image analysis in gastroesophageal (GE) adenocarcinomas in \cite{jeung12}. They achieved an overall correlation of 84\%. 

HER2-CONNECT presented a 92.3\% agreement between the HER2-CONNECT software and the pathologists according to \cite{brugmann12}. This software exploits the ability of computer image analysis to quantify the standard HER-2 IHC ”wire mesh” pattern by measuring the connectivity and size distribution of colored membranes. Their approach is based on brown segmentation, membrane skeletonization and elimination of noise. HER-2 score is defined based on the size of the membrane distribution and the area it occupies.

A comparison of slidePath's tissue IA system with other commercially available systems for HER-2-analysis are presented in the study conducted by \cite{dobson10}, which determined HER-2 score as 0/1+, 2+ or 3+ (negative, limit and positive). The concordance with manual review are: Slidepath: 91\%, Aperio: 86\%, BioImagene: 81\%, Dako (Chromavision): 75\% and Ventana (TriPath Imaging): 86\% and 77\%. 

A limitation of commercial systems is that they require manual intervention, in the sense that they are trained for a particular biomarker set and need to be manually optimized. Such adjustments introduce subjective criteria and become sources of inter-laboratory variability \cite{masmoudi09}. Since the systems have these limitations and are expensive, alternatives to this problem are still been developed. Also, a segmentation step is generally necessary.

The method described in \cite{masmoudi09} is a multi-stage algorithm, with an agreement of 81\%-83\%. The algorithm steps are color pixel classification, nuclei segmentation, and cell membrane modeling, and extracts quantitative, continuous measures of cell membrane staining intensity and completeness. A minimum cluster distance classifier merges the features to classify the slides into HER-2 categories.

The study presented in \cite{hall08} used color decomposition based on polar transform, threshold and gaussian filters, resulting in an AUC (Area Under the Curve) of 87\%. A correlation of 84\% was obtained in \cite{joshi07} by preprocessing image and RGB channels segmentation. More recently, a study about deep learning applied on this topic obtained a concordance of 83\% with a pathologist \cite{vandenberghe17}. They used ConvNets for segmentation, feature extraction and classification techniques for cell/nucleus detection. In \cite{gaur2016membrane}, a transfer learning mechanism based on active learning was applied to segment membrane in FISH images.

\cite{saha18} developed a deep learning framework for detection, segmentation and classification of cell membranes and nuclei from HER-2 stained breast cancer images, achieving 98.33\% accuracy. The proposed method was assessed based on the HER-2 challenge contest image database of University of Warwick \cite{qaiser18}.

This challenge received 18 submissions, which most applied a supervised patch-based classification approach to handle the problem. A common pipeline was based on three main components: 1) preprocessing, including identification of regions of interest, 2) patch classification based on handcrafted or neural network learned features and 3) techniques to define the HER-2 score at WSI (Whole Slide Image) level.
The best result on this competition built a handcrafted sub-dataset. For this purpose, a set of 68x68 patches was extracted from training. GoogLeNet and a percent-based rule were used for HER-2 score classification.

Aiming to bring a method without manual intervention and segmentation, we propose a fully-automated classification based on color features, thus reducing the complexity in this analysis. Section \ref{sec:methods} describes the dataset and methods applied in the classification. Experimental results are reported in Section \ref{sec:results}, then proposing future works and a conclusion in Section \ref{sec:conclusion}.

\section{Materials and Methods} \label{sec:methods}

\subsection{Dataset}
\label{ssec:dataset}

The proposed method was developed based on the HER-2 image database of the Department of Computer Science, University of Warwick, United Kingdom \cite{qaiser18}. The dataset entailed 172 whole slide images (WSI) extracted from 86 cases of invasive breast carcinomas and included both the H\&E (Haematoxylin \& Eosin) and HER-2 stained slides. Images stained with H\&E are used in routine diagnostic practice of BC to identify tumour regions. Our approach only uses the HER-2 stained slides, being composed of 52 images for training and 34 for testing.

\begin{figure}[ht]
\centering
\includegraphics[width=.7\textwidth]{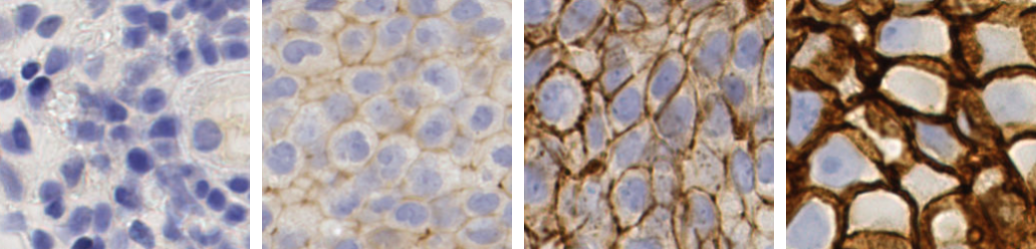}
\caption{Example of classes 0, 1+, 2+, 3+}
\label{fig:classes}
\end{figure}

The histology slides for this contest were scanned on a Hamamatsu NanoZoomer C9600, enabling the image to be viewed from a $\times$4 to a $\times$40 magnification. Each WSI was cropped in patches, at $\times$40 magnification, by a OpenSlide \cite{goode13} function, each one in size of 250x250 pixels. The patches with more tissue information were automatically selected by analyzing their histogram. Figure \ref{fig:classes} illustrates examples of classes' patches.

The authors of this dataset only provided ground truth for training images. It is required to submit the algorithm to evaluate on testing images. Therefore, we present on this paper only evaluation in the training subset. 
We might report test evaluation in future works. 

\subsection{Proposed approach}
\label{sec:proposed_approach}
\begin{figure}[ht]
\centering
\includegraphics[width=.99\textwidth]{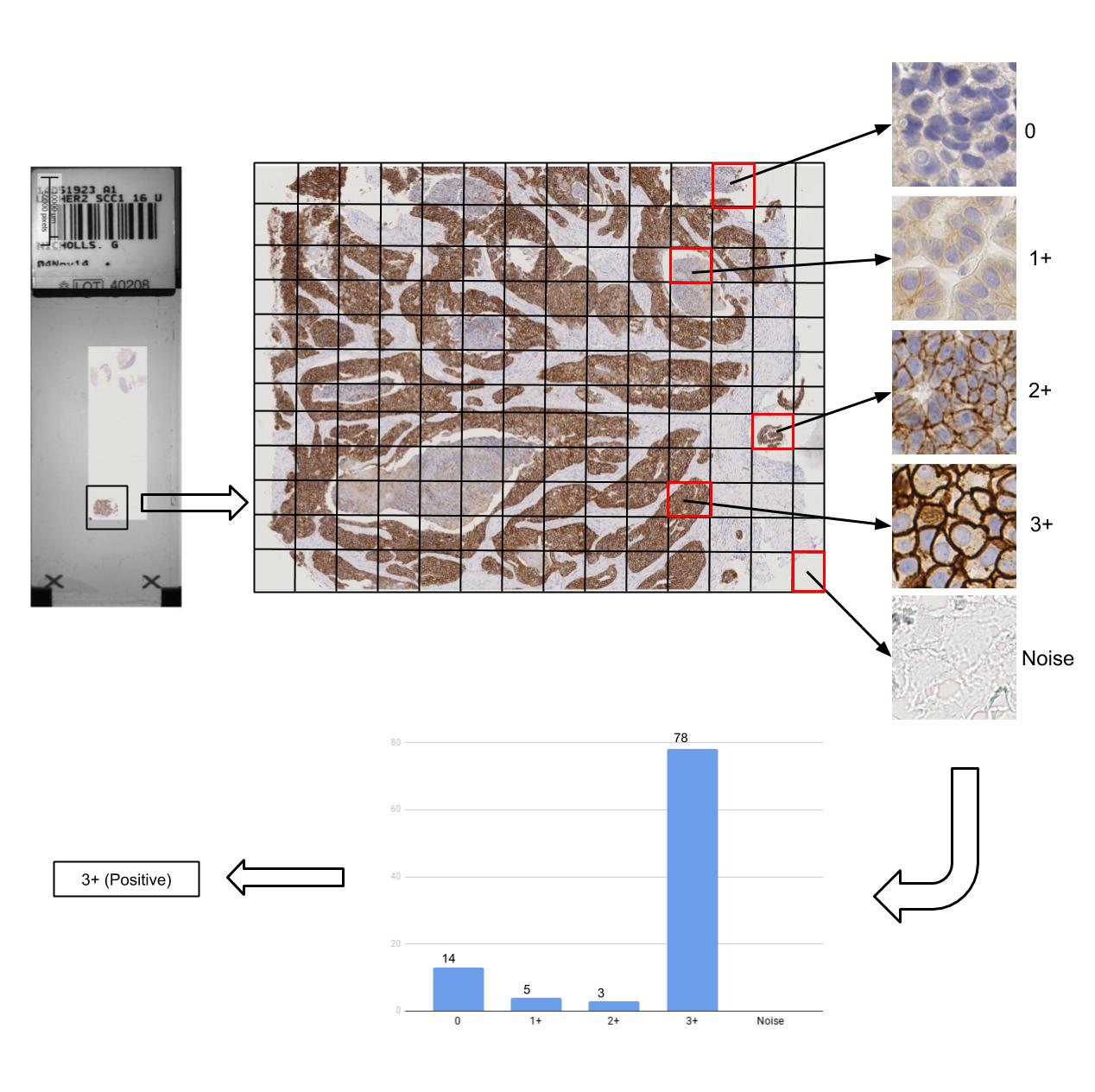}
\caption{An illustration of our method}
\label{fig:method}
\end{figure}

Our approach is divided into two levels: image and patient. In the first level, we analysis the classification of individual patches. Then, based on the analysis done in the previous level, occurrence of each class' patches is employed to predict HER-2 score. Figure \ref{fig:method} illustrates our algorithm pipeline.

\textbf{Image level}: The purpose of this step is to state the features which best represent the relevant patches. Assisted by a pathologist, the most representative patches (generally around 30) were selected out of each WSI. This amount was decided to balance relevant ones among total patches of each slide. Firstly, color and texture features were extracted. For color features, histograms in RGB and HSV models, additionally with mean and standard deviation of each channel, were experiment. For texture, we employed the LBP (Local Binary Pattern) \cite{ojala96} and PFTAS (Parameter-free Threshold Adjacency Statistic) \cite{coelho10} descriptors. Then, SVM (Support Vector Machine) \cite{vapnik95}, KNN (K-Nearest Neighbor) \cite{belur91}, MLP (Multilayer Perceptron) \cite{LeCun098} and Decision Tree \cite{breiman84} classifiers' accuracy were evaluated using leave-one-patient-out validation. In this step, we also trained classifiers to distinguish noise patches.

\textbf{Patient level}: The best descriptors in the image level were used to classify all patches in each exam. Although a WSI is scored in only one class, these slides may have patches from different classes. Therefore, we need to set a rule for HER-2 scoring. A threshold rule based on the quantity of patches from each class was experimented, but results were not satisfactory. Then, we used each class occurrence as input for a classifier, creating a feature vector of occurrences. Classification is then applied to determine the HER-2 score of the WSI. The same classifiers from the previous step were employed, also accuracy using leave-one-patient-out validation were implemented. 

Since clinical decisions do not differentiate 0 and 1+ classes and only consider tests as negative (0/1+), limit (2+) and positive (3+) \cite{tuominen12}, we have developed two approaches: with four (0, 1+, 2+, 3+ and noise) and five classes (negative, limit, positive and noise). Our method evaluates classifiers' accuracy using leave-one-patient-out validation and basically consists in: 
\begin{enumerate}
   \item Crop a WSI in patches of size 250x250;
   \item Classify each patch using training patches selected by a pathologist (image level);
   \item Create a vector with percentage of patches from each class;
   \item Test classifiers to define HER-2 score based on these percentages (patient level).
\end{enumerate}

\subsubsection{Descriptors and Classifiers Parameters}
\label{sec:parameters}

To clarify the experiments, this section presents the parameters used in descriptors and classifiers algorithms.
We extracted LBP features using non-rotation-invariant uniform patterns variant, $radius=1$ and 8 neighbours. PFTAS was implemented using mahotas function \cite{coelho13}.
GridSearch was applied to find the best parameters for SVM. A exhaustively search for $c$, $gamma$ and $kernel$, parameters of the classifier, is performed for this function. Best values were employed in each experiment.
Euclidean Distance was calculated in KNN. Variations of $k$, from 1 to 9, were analyzed for KNN, where the best results were obtained with $k=1$.
MLP and Decision Trees were implemented with defaults parameters in scikit-learn library \cite{pedregosa11}.
These methods remain to be more explored.

In the next section, we present the results obtained using the proposed approach.

\section{Experimental Results} \label{sec:results}

Firstly, the results in image level are shown. Table \ref{tab:train_accuracy} presents the accuracy resulted of leave-one-patient-out validation on training patches (those selected by a pathologist).

\begin{table}[ht]
\centering
\caption{Accuracy on image level - training patches (in \%)}
\label{tab:train_accuracy}
\begin{tabular}{lccccccccc}
\toprule 
& \multicolumn{4}{c}{(0/1+), 2+, 3+ and NOISE} & & \multicolumn{4}{c}{0, 1+, 2+, 3+ and NOISE} \\
& \textbf{SVM} & \textbf{KNN} & \textbf{MLP} & \textbf{Tree} & & \textbf{SVM} & \textbf{KNN} & \textbf{MLP} & \textbf{Tree} \\ \midrule
\textbf{HSV}            & 88.44 & 87.75 & \textbf{90.20} & 84.09 & & 82.67 & 80.60 & \textbf{86.51} & 82.52 \\
\textbf{HSV\_MS} & 88.62 & 87.80 & \textbf{89.70} & 85.87 & & 82.77 & 81.07 & \textbf{85.22} & 83.14 \\
\textbf{HSV\_RGB}       & 88.36 & 86.16 & \textbf{89.40} & 85.18 & & 82.55 & 76.44 & \textbf{85.71} & 81.76 \\
\textbf{LBP}            & \textbf{58.37} & 49.83 & 57.67 & 49.82 & & \textbf{50.80} & 41.83 & 49.37 & 39.87 \\
\textbf{PFTAS}          & \textbf{79.87} & 68.99 & 76.03 & 69.64 & & \textbf{69.77} & 60.27 & 68.44 & 59.53 \\ \bottomrule
\end{tabular}
\end{table}

Analyzing our results, the texture descriptors employed did not discriminate the evaluated patches correctly. 
We obtained satisfactory results in both approaches, with four and five classes.
Then, only color descriptors were used in patient level classification. 
Color descriptors with SVM, KNN ($k=1$), MLP and Decision Tree were used in image level, to distinguish patches and create probabilities to scoring HER-2. By using these probabilities to predict HER-2 score, SVM, KNN, MLP and Decision Tree were experimented in patient level. 

Regarding the performance at patient level, Table \ref{tab:her2_accuracy} shows an overall increase in accuracy when classifying only with three classes, probably related to the similarity among 0 and 1+ classes. Also, SVM is a promising classifier to scoring HER-2 based on probabilities created by any color descriptor which classified patches using KNN classifier(HSV+KNN, HSV\_MS+KNN, HSV\_RGB+KNN). 

\begin{table}[ht]
\centering
\caption{Accuracy on patient level - HER-2 scoring (in \%)}
\label{tab:her2_accuracy}
\begin{tabular}{llllllllll}
\toprule
& \multicolumn{4}{c}{(0/1+), 2+ and 3+} & & \multicolumn{4}{c}{0, 1+, 2+, 3+} \\
& \textbf{SVM} & \textbf{KNN} & \textbf{MLP} & \textbf{Tree} & & \textbf{SVM} & \textbf{KNN} & \textbf{MLP} & \textbf{Tree} \\ \midrule
\textbf{HSV+SVM}             & 86.27 & \textbf{90.20} & 84.31 & 88.24 & & 61.54 & \textbf{65.38} & 50.00 & 61.54 \\ 
\textbf{HSV+KNN}             & \textbf{94.12} & 86.27 & 92.16 & 92.16 & & \textbf{86.27} & 70.59 & 78.43 & 72.55 \\ 
\textbf{HSV+MLP}             & \textbf{86.27} & 82.35 & 86.27 & 76.47 & & \textbf{67.31} & 65.38 & 51.92 & 50.00 \\ 
\textbf{HSV+Tree}            & \textbf{78.43} & 60.78 & 70.59 & 78.43 & & \textbf{53.85} & 30.77 & 46.16 & 48.08 \\ \midrule
\textbf{HSV\_MS+SVM}  & \textbf{86.27} & 86.27 & 86.27 & 76.47 & & 57.69 & 59.62 & 53.85 & \textbf{65.38} \\ 
\textbf{HSV\_MS+KNN}             & \textbf{94.12} & 86.27 & 92.16 & 88.24 & & \textbf{86.27} & 70.59 & 74.51 & 74.51 \\ 
\textbf{HSV\_MS+MLP}  & 86.27 & 84.31 & \textbf{88.24} & 84.31 & & \textbf{73.08} & 51.92 & 57.69 & 67.69 \\ 
\textbf{HSV\_MS+Tree} & 68.63 & 62.75 & \textbf{72.55} & 70.59 & & \textbf{59.62} & 59.62 & 50.00 & 46.16 \\ \midrule
\textbf{HSV\_RGB+SVM}        & \textbf{86.27} & 80.39 & 86.27 & 86.27 & & 55.77 & 55.77 & 50.00 & \textbf{57.69} \\ 
\textbf{HSV\_RGB+KNN}             & \textbf{94.12} & 82.35 & 86.27 & 84.31 & & \textbf{84.31} & 70.59 & 74.83 & 72.55 \\ 
\textbf{HSV\_RGB+MLP}        & \textbf{88.24} & 82.35 & 86.27 & 76.47 & & \textbf{73.08} & 55.77 & 63.46 & 67.31 \\ 
\textbf{HSV\_RGB+Tree}       & \textbf{84.31} & 66.67 & 82.35 & 62.75 & & \textbf{61.54} & 40.38 & 50.00 & 46.15 \\ \bottomrule
\end{tabular}
\end{table}

Although SVM and MLP had better results in the image level evaluation, the feature vector generated from classes' occurrences by the KNN achieved an overall higher accuracy. A likely motive is that the two classifiers did not adapt to patches outside the ones selected by the pathologist, which were more class homogeneous. Since patches can be heterogeneous, meaning that each patch has certain cells and membrane which can be classified in different HER-2 scores, it is difficult to correctly represent them. KNN seems to better classify these peculiarities. Due to this heterogeneity other probability descriptors created by SVM, MLP and Tree were not discriminative for HER-2 score.

The worst results in patient level were resulted from patches classified by Decision Tree with any descriptor (HSV+Tree, HSV\_MS+Tree, HSV\_RGB+Tree). As it is observed in Table \ref{tab:train_accuracy}, this classifier did not perform well in image level. Despite the fact the accuracy percentage is not much lower than others, the results presented in this step only involve patches analyzed by the pathologist. In a WSI, more difficult patches can be present, with more heterogeneous classes and thus, this result appears to be a consequence of the image level classification.

Tables \ref{tab:mat_conf} shows a confusion matrix of the best results obtained (three classes classification). In our approach we consider 0/1+ as negative, 2+ as limit and 3+ as positive. 

\begin{table}[ht]
\centering
\caption{SVM Confusion Matrix}
\label{tab:mat_conf}
\resizebox{0.7\textwidth}{!}{
\begin{tabular}{rccccccccccc}
\toprule
              & \multicolumn{3}{c}{HSV+KNN}      &  & \multicolumn{3}{c}{HSV\_MS+KNN}  &  & \multicolumn{3}{c}{HSV\_RGB+KNN} \\ \midrule
              & \textbf{0/1+} & \textbf{2+} & \textbf{3+} &  & \textbf{0/1+} & \textbf{2+} & \textbf{3+} &  & \textbf{0/1+} & \textbf{2+} & \textbf{3+} \\ \midrule
\textbf{0/1+} & 24            & 0           & 0           &  & 23            & 1           & 0           &  & 23            & 1           & 0           \\
\textbf{2+}   & 2             & 11          & 1           &  & 1             & 12          & 1           &  & 1             & 12          & 1           \\
\textbf{3+}   & 0             & 0           & 13          &  & 0             & 0           & 13          &  & 0             & 0           & 13         \\ \bottomrule
\end{tabular}}
\end{table}

In CADs (Computer Aided Decision) focused on cancer treatment decision, it is important to evaluate specificity and sensitivity. The three confusion matrices presented 100\% sensitivity and specificity. It means patients that should receive treatment with trastuzumab will receive it and patients that do not need to be treated with trastuzumab, will not be. These metrics are about negative(0/1+) and positive classes(3+). Mistakes between 2+ and other are very common, thus a FISH test is required to confirm HER-2 positivity in 2+ slides. Despite some 2+ confusions, our results are still very promising and might assist pathologists as a second opinion.

Our method avoids segmentation and do not need manual intervention, different of several works reviewed. In Table \ref{tab:comparision} we compared our method with others described before. HER2NET proposed by \cite{saha18} had a better accuracy than ours. Although both works have used the same dataset, partitions for training and test were different. Also, HER2NET depends on manual intervention for ROI selection and includes a segmentation step.

\begin{table}[ht]
\centering
\caption{Comparison with related works}
\label{tab:comparision}
\resizebox{0.9\textwidth}{!}{
\begin{tabular}{lccc}
\toprule
                            & \textbf{Manual Intervention}   & \textbf{Segmentation}     & \textbf{Remarks}\\ \midrule
\cite{brugmann12}           & Yes                           & Yes                       & 92.3\% agreement\\
\cite{masmoudi09}           & Yes                           & Yes                       & 83\% agreement\\
\cite{hall08}               & Yes                           & Yes                       & 87\% AUC\\
\cite{joshi07}              & Yes                           & Yes                       & 84\% correlation\\
\cite{vandenberghe17}       & No                            & Yes                       & 83\% concordance\\
\cite{saha18}               & Yes                           & Yes                       & 98.33\% accuracy\\ 
\textbf{Proposed work}      & \textbf{No}                   & \textbf{No}               & \textbf{94.12\% accuracy}\\ \bottomrule
\end{tabular}}
\end{table}

\section{Conclusion} \label{sec:conclusion}

The purpose of this study is to provide a technique able to scoring HER-2 in histopathological slides. Our results show that the proposed approach using classical machine learning techniques and color descriptors is very promising. Since we only used simple features and also without combining them, results may be improved by a more broadly study of descriptors and combination of them. Also, we have a limitation about the number of samples. Studies in other datasets and with a greater volume of samples may lead to improvements and show a more reliable result.

As described in literature review, most classical approaches include segmentation, which is known to introduce errors in next steps. Their concordance was around 85\%, being increased by using deep learning techniques. Nonetheless, our approach achieved more than 90\% accuracy, avoiding explicit segmentation and extraction of structure properties such as cell nuclei, membrane, size and shape of these. Besides, it is fully automated and can easily works in simple desktop computers. Thus, findings presented in this study support the idea of cheap techniques to help in pathologists routine.

Furthermore, we propose to compare classical machine learning and deep learning techniques, and also to employ images obtained in different clinical conditions.

\bibliographystyle{sbc}
\bibliography{sbc-template}

\end{document}